%% file: main.tex
\documentclass[10pt,twocolumn,letterpaper]{article}

\usepackage{cvpr}              

\input{preamble}

\usepackage{bm}
\definecolor{cvprblue}{rgb}{0.21,0.49,0.74}
\usepackage[pagebackref,breaklinks,colorlinks,citecolor=cvprblue]{hyperref}
\usepackage{tabularx}
\def\paperID{6004} 
\def\confName{CVPR}
\def\confYear{2024}

\title{How to Configure Good In-Context Sequence for Visual Question Answering}

\author{\\Li Li$^{1}$ \quad Jiawei Peng$^1$ \quad Huiyi Chen$^1$\quad Chongyang Gao$^2$\quad Xu Yang$^1$\footnotemark[1]\\
\normalsize$^1$ School of Computer Science \& Engineering, Key Lab of New Generation Artificial Intelligence\\
\normalsize Technology \& Its Interdisciplinary Applications (Ministry of Education), Southeast University\\
\normalsize$^2$ Department of Computer Science, Northwestern University\\
\tt\small lilyli@seu.edu.cn, pengjiawei@seu.edu.cn, huiyichen@seu.edu.cn,\\ \tt\small chongyanggao2026@u.northwestern.edu, xuyang\_palm@seu.edu.cn \\
}

\begin{document}
\setlength{\abovedisplayskip}{2pt}
\setlength{\belowdisplayskip}{2pt}
\maketitle
\renewcommand{\thefootnote}{\fnsymbol{footnote}}
\footnotetext[1]{Corresponding author.}

\input{sec/0_abstract}    
\input{sec/1_intro}
\input{sec/2_related}
\input{sec/3_approach}
\input{sec/4_experiment}
\input{sec/5_conclusion_limitations}
{
    \small
    \bibliographystyle{ieeenat_fullname}
    \bibliography{main}
}
\appendix
\input{sec/6_appendix}

\end{document}

%% file: preamble.tex
\usepackage[dvipsnames]{xcolor}


%% file: sec/0_abstract.tex

\begin{abstract}
Inspired by the success of Large Language Models in dealing with new tasks via In-Context Learning (ICL) in NLP, researchers have also developed Large Vision-Language Models (LVLMs) with ICL capabilities. However, when implementing ICL using these LVLMs, researchers usually resort to the simplest way like random sampling to configure the in-context sequence, thus leading to sub-optimal results. To enhance the ICL performance, in this study, we use  Visual Question Answering (VQA) as case study to explore diverse in-context configurations to find the powerful ones. Additionally, through observing the changes of the LVLM outputs by altering the in-context sequence, we gain insights into the inner properties of LVLMs, improving our understanding of them. Specifically, to explore in-context configurations, we design diverse retrieval methods and employ different strategies to manipulate the retrieved demonstrations. Through exhaustive experiments on three VQA datasets: VQAv2, VizWiz, and OK-VQA, we uncover three important inner properties of the applied LVLM and demonstrate which strategies can consistently improve the ICL VQA performance. Our code is provided in: \url{https://github.com/GaryJiajia/OFv2_ICL_VQA}.
\end{abstract}

%% file: sec/1_intro.tex
\section{Introduction}
\label{sec:intro}

\begin{figure*}[htb]
  \centering
  \includegraphics[width=16cm]{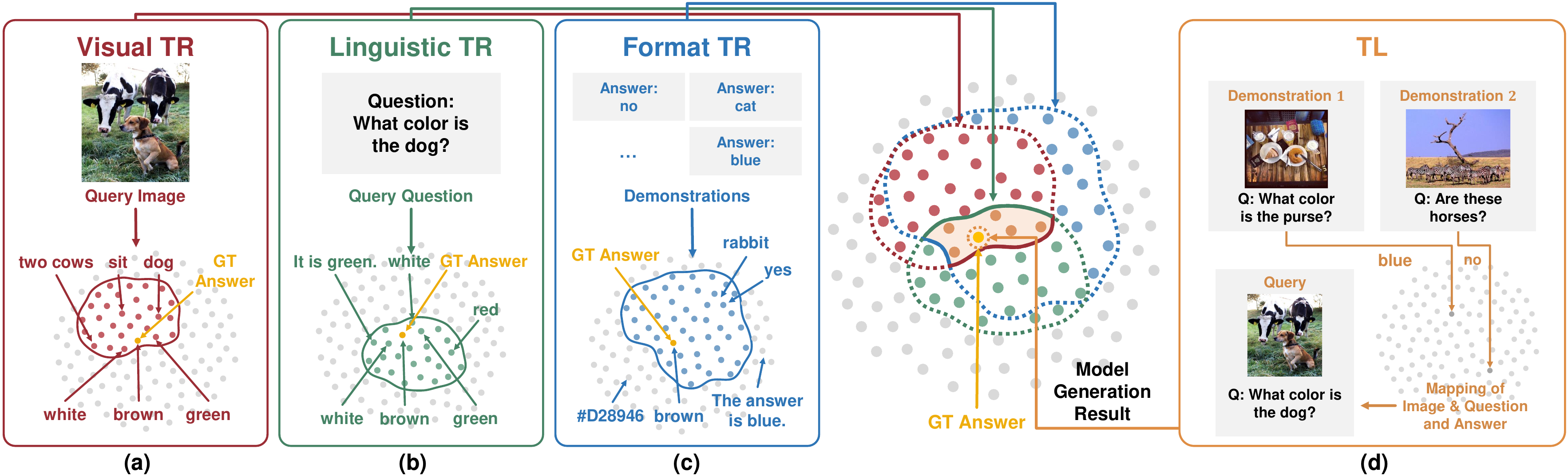}
  \vspace{-0.05in}
  \caption{ICL shows two different functions: Task Recognition (TR) and Task Learning (TL). In VQA, TR includes three components: (a) Visual TR and (b) Linguistic TR narrow the label space based on the query image and question, and (c) Format TR recognizes the answer formats from the demonstrations. After combining them, the label space can be narrowed down for better answer the question. While TL learns the mapping between inputs (images\&questions) and outputs (answers) from demonstrations to make LVLM get the right answer.}
  \label{fig:tl&tr}
  \vspace{-0.15in}
\end{figure*}

Recently, Large Language Models (LLMs)~\cite{devlin2018bert,radford2019language,brown2020language} have showed remarkable abilities in solving new tasks through prompt engineering~\cite{liu2023prompt} and In-Context Learning (ICL)~\cite{dong2022iclsurvey}. 
However, despite their success, LLMs still remain inscrutable to the research community.
To unravel the properties of these models, researchers have drawn inspiration from the ``outside-in" methodologies to comprehend complex systems.
Analogously, as scientists treat unknown systems as black boxes, conducting experiments to discern the effects of varied inputs on outputs, researchers introduce diverse prompts and analyze the resultant feedback. This strategy provides crucial insights into the inner properties of LLMs\cite{min2022rethinking,tr&tl}.

Compared to the standard single sentence prompt, which is one kind of zero-shot prompt, ICL sequences few-shot \textbf{demonstrations} where each one contains knowledge about the input and the corresponding label of the task that needs to be solved. Such few-shot nature of ICL enables it to encapsulate more information, resulting in enhanced performance. 
However, the ICL performance is heavily influenced by various demonstration configurations, such as the selection or ordering of the demonstrations~\cite{liu2021selection,kim2022self,lu2021order}. Consequently, many NLP studies~\cite{rubin2021nlpretrieve,zhang2022nlpretrieve,wu2022nlpretrieve,gonen2022nlpretrieve} explore how to configure demonstrations to enhance the ICL performance.
Meanwhile, NLP researchers also use ICL to unravel the inner properties of large models, owing to its flexible controllability. For example, by controlling the label space of the demonstrations, researchers~\cite{tr&tl} find that the ICL ability may be demonstrated by two distinct functions: Task Recognition (TR), \ie, the ability to identify the task formulation, and Task Learning (TL), the ability to learn the mapping between input and labels of the demonstrations.

Inspired by the success of LLM, researchers in the vision-language (VL) domain have also developed large models with ICL capabilities, such as Flamingo~\cite{alayrac2022flamingo} and its corresponding open-source version, Open-Flamingo~\cite{awadalla2023openflamingo}. However, there is limited research on how to effectively configure demonstrations in these models, both in terms of enhancing the performance of Large Vision-Language Model (LVLM) and exploring its properties. To the best of our knowledge, currently, only one study \cite{wylnips} has explored demonstration configurations for image captioning. Unfortunately, this research still fails to make use of ICL to explore the properties of LVLM. 

In VL, Visual Question Answering (VQA) 
is more suitable for exploring the inner properties of LVLM through ICL for two reasons. 
First, most NLP tasks employed to explore LLMs can be considered as question-answering tasks, \ie, the sentiment classification can be viewed as answering ``what is the sentiment of this sentence?". Therefore, VQA is well-suited for adapting the methodologies used in QA studies into the VL domain. Second, VQA encompasses various visual understanding tasks, including classification, counting, locating and so on, allowing for a more comprehensive exploration of LVLM. Therefore, in this study, we explore demonstration configurations in the VQA task with a dual-purpose: \textit{(1) to explore effective demonstration configuration strategies for enhancing VQA performance} and \textit{(2) to gain a better understanding of the inner properties of LVLM}.

To achieve the dual-purpose, we design various demonstration configuration strategies, including retrieving demonstrations based on similarity via images or texts (questions and answers) and using different ways to manipulate the in-context sequence constructed by the retrieved demonstrations, \eg, mismatching the (image, question, answer) triplets, incorporating the instructions, and reordering the demonstrations. Through exhaustive experiments, our research makes the following three key contributions.

\begin{itemize}
\item We extend the TR and TL hypothesis to the field of LVLM by refining this hypothesis to interpret and measure the ICL capabilities of LVLM, as depicted in \cref{fig:tl&tr}.
\item Based on the refined hypothesis, we uncover three important inner properties of LVLM during ICL: limited TL capabilities, the presence of a short-cut effect, and partial compatibility between vision and language modules.
\item Building upon these findings, we explain in detail the roles played by various demonstration configuration strategies in LVLM and design new demonstration configuration methods.
\end{itemize}

%% file: sec/2_related.tex
\section{Related Work}
\label{sec:related_work}

\noindent\textbf{In-Context Learning in NLP.}
Recently, NLP has witnessed significant advancements in Large Language Models (LLMs).
With the increase in model and corpus sizes~\cite{devlin2018bert,radford2019gpt2}, researchers discovered their emergent capabilities, particularly in prompt engineering~\cite{liu2023promptsurvey,li2021prefix,lester2021prompttuning,liu2023prompt}. The introduction of even larger models like GPT-3~\cite{brown2020gpt3} has unveiled the potential for In-Context Learning (ICL).
ICL, a form of specialized prompt engineering, enables LLMs to make predictions based on contextual information supplemented by a few illustrative examples. Numerous investigations have demonstrated the proficient performance of LLMs in various tasks through ICL~\cite{wei2022emergent,wei2022CoT}.
This led to a surge of studies exploring the configuration of in-context sequence~\cite{gao2020nlpicl,jiang2020nlpicl,su2022nlpicl,levy2022diverse,lu2021order}. However, most of these studies have been limited to NLP tasks, and there is a need to extend this research to other domains.

\noindent\textbf{In-Context Learning in VL.}
Inspired by the success of LLMs in NLP, the vision-language field has also witnessed the emergence of corresponding large vision-language models (LVLMs)~\cite{lu2019vilbert,chen2020uniter,li2020oscar}. Some of these models, such as BLIP2~\cite{li2023blip2}, MiniGPT-4~\cite{zhu2023minigpt4}, and LLAVA~\cite{liu2023llava}, are pretrained by aligning image and text data using adapters~\cite{li2022blip,yu2022coca} to alleviate training burdens. Specifically, they freeze a well-trained LLM and train a smaller network alongside it, leveraging this alignment to enable joint learning from both modalities during pretraining.
Although there are numerous large VLMs available, it is important to note that not all models support in-context learning (ICL). For example, mPLUG-Owl~\cite{ye2023mplugowl} and MiniGPT-4~\cite{zhu2023minigpt4} lack the capabilities for ICL because they have not undergone dedicated few-shot pre-training and cannot handle the input distribution associated with in-context learning. In contrast, models like Flamingo~\cite{alayrac2022flamingo} and Otter~\cite{li2023otter} are specifically designed to support this task. 
However, since Flamingo is not open-source, we utilize its open-source version called Open-Flamingo~\cite{awadalla2023openflamingo}. 
Among them, Otter derives from Open-Flamingo through instruction fine-tuning. In our research, we utilize Open-Flamingo, removing the interference caused by instruction fine-tuning. Additionally, the recently released MMICL~\cite{zhao2023mmicl} model includes pre-training data from the classic VQA datasets such as VQAv2~\cite{goyal2017vqav2}, VizWiz~\cite{bigham2010vizwiz}, and OK-VQA~\cite{marino2019okvqa}. Open-Flamingo, on the other hand, does not use these datasets for pre-training, thus eliminating any interference from being exposed to them during the pre-training process.
Therefore, Open-Flamingo emerges as the most suitable choice for conducting ICL research at present.
We employ two versions of Open-Flamingo, namely Open-Flamingo v1 and v2.

Currently, there is limited research on multimodal ICL, with only one study focusing on captioning~\cite{wylnips}. We are the first to explore demonstration configuration in the context of the Visual Question Answering (VQA) task.

\noindent\textbf{Configuring In-Context Sequence for QA.}
In NLP, there is a significant body of research dedicated to demonstration configuration. This research encompasses techniques such as leveraging similarity measures to retrieve more relevant in-context examples~\cite{liu2021selection} or employing machine-generated demonstrations~\cite{kim2022self}. During the research process, some studies have also identified certain properties of LLMs when applied to in-context learning. For instance, ~\cite{min2022rethinking} discovered that randomly replacing labels in demonstrations has minimal impact on performance, and as long as the demonstration maintains consistency in terms of format, input distribution, label space, and query, the model can achieve favorable results. 
~\cite{lu2021order} have empirically demonstrated that order sensitivity is a common and persistent challenge across various models. 
Additionally, ~\cite{tr&tl} proposed a deconstruction of ICL into task recognition and task learning, investigating the TR and TL capabilities of models with different shot numbers and scales.
Furthermore, ~\cite{lyu2022copyingeffect} observed the presence of a ``copying effect" phenomenon within LLMs, which is a type of short-cut inference.

%% file: sec/3_approach.tex
\section{In-Context Learning (ICL) for VQA}

Given a well-trained Large Vision-Language Model (LVLM) \eg, Flamingo~\cite{alayrac2022flamingo}, we can use it to solve VQA by ICL. To achieve this, we need to prepare a multi-modal in-context sequence $\mathcal{S}=\{(\bm{I}_1,\bm{Q}_1,\bm{A}_1);(\bm{I}_2,\bm{Q}_2,\bm{A}_2);...;(\bm{I}_n,\bm{Q}_n,\bm{A}_n);(\hat{\bm{I}},\hat{\bm{Q}})\}$ that consists of $n$-shot (image $\bm{I}$, question $\bm{Q}$, answer $\bm{A}$) triplets acting as the demonstrations and one test sample $(\hat{\bm{I}},\hat{\bm{Q}})$. Then we input $\mathcal{S}$ to the LVLM for generating the corresponding answer $\hat{\bm{A}}=\{\hat{a}_1,...,\hat{a}_T\}$, where the $t$-th word $\hat{a}_t$ is sampled from the following word distribution $P(\cdot)$ calculated by the LVLM:
\begin{equation} \label{equ:icl}
    P(\hat{a}_t|\mathcal{S},\hat{a}_{1:t-1})
\end{equation}

Next, we first extend the hypothesis that ICL can be decoupled into Task Recognition (TR) and Task Learning (TL) from NLP~\cite{tr&tl} to VL domain in \cref{sec:approach_tr&tl} where we further decouple TR into format TR, visual TR, and linguistic TR for better analyzing the ICL ability of a LVLM. Then we introduce the techniques used to configure the demonstrations. The applied techniques include two parts where the first part in \cref{sec:retrieve_demonstrations} shows how to retrieve the samples from a supporting set and the second part in \cref{sec:manipulate_demonstrations} discusses how to manipulate the in-context sequence constructed by the retrieved demonstrations. Due to space constraints, only the effective techniques we utilize are presented here and some other less effective ones are given in the supplementary materials.

\subsection{TR and TL in the VL domain\label{sec:approach_tr&tl}}
The ability of ICL can be demonstrated by two distinctive functions: Task Recognition (TR) and Task Learning (TL)~\cite{tr&tl}. TR recognizes the task based on the demonstrations, \eg, recognizing the data distribution of the task, and applying pre-trained priors of LLM. While TL focuses on learning the correct input-output mapping from the demonstrations, which can be regarded as an implicit learning process analogous to explicit fine-tuning ~\cite{dai2023implicitgradient}.

In this paper, we further refine this hypothesis, providing a more detailed interpretation within the VL realm. Specifically, we further decouple TR into three aspects: format TR, visual TR, and linguistic TR, as shown in \cref{fig:tl&tr}. Format TR pertains to the capacity of the LVLM to identify the task format, input distribution, and label space based on demonstrations. For example, in \cref{fig:tl&tr} (c), the question-answer format of the demonstration helps the model determine that the potential answer should be a single word or a simple phrase rather than a complete sentence. 

Visual and linguistic TR correspond to the recall of corresponding pre-trained knowledge preserved in the LVLM. As \cref{fig:tl&tr} (a) shows, visual TR uses a visual encoder to identify relevant labels associated with the query image, including the appeared entities, colors, relationships, and more. Linguistic TR (\cref{fig:tl&tr} (b)) recognizes the query question (e.g., ``What color is the dog?") through the language component. Drawing from pre-training experience, the model delimits the potential answer space, indicating that only labels related to colors are admissible as the answers. By combining three TR abilities, the label space can be narrowed down for LVLM to make better prediction. 

On the other hand, TL refers to the ability of the LVLM model to learn the mapping relationship between (image, question) pairs and their corresponding answers from the demonstrations. As shown in \cref{fig:tl&tr} (d), TL treats the questions and ground truth answers from demonstrations as ``training samples", from which it learns the mapping. Then if the LVLM can successfully achieve TL, it can directly map the query into the correct answer.


\begin{figure*}[htb]
  \centering
  \includegraphics[width=16cm]{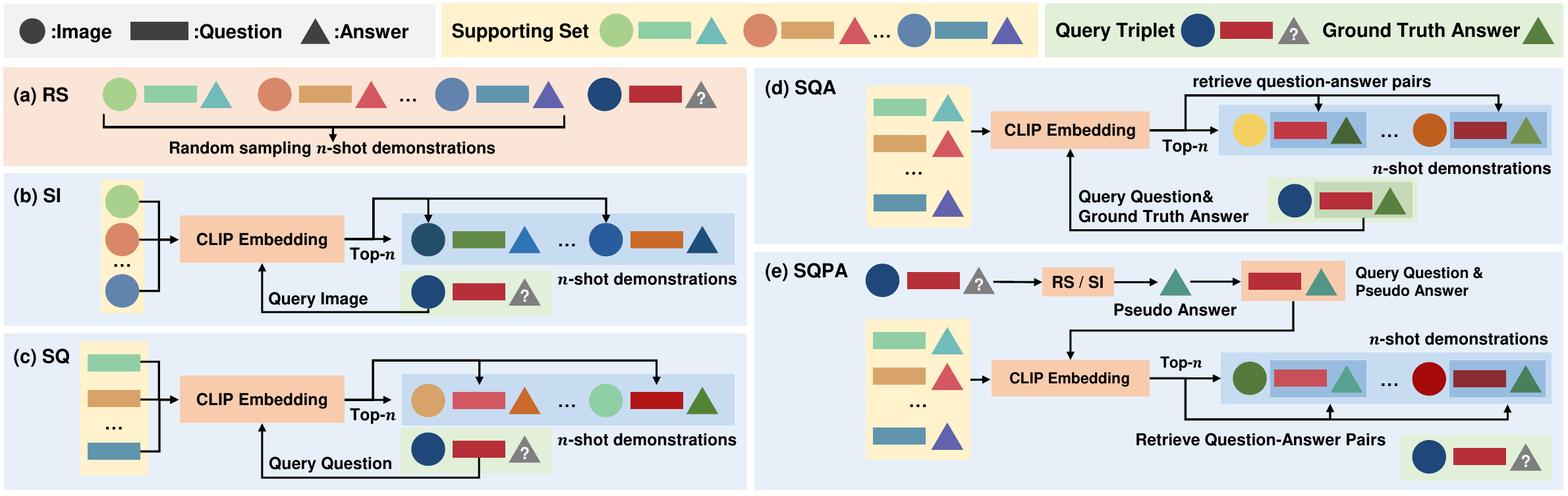}
  \vspace{-0.05in}
  \caption{The schematic representation of the demonstrations retrieval strategies. Circles, rectangles, and triangles respectively represent the images, questions, and answers in the triplet. The color proximity between these elements indicates their similarity level.}
  \label{fig:retrieve}
  \vspace{-0.15in}
\end{figure*}

\subsection{Retrieving Demonstrations}
\label{sec:retrieve_demonstrations}

Recognizing that each component (\eg, image, question, and answer) of a VQA sample can be used as an index, we can respectively use them to retrieve $n$ examples from the supporting set $\mathcal{D}=\{(\bm{I}_1,\bm{Q}_1,\bm{A}_1),...,(\bm{I}_N,\bm{Q}_N,\bm{A}_N))\}$ as the demonstrations for $n$-shot setting. After that, we can sequence the $n$-shot triplets to construct the in-context sequence $\mathcal{S}$. Next we introduce specific retrieval strategies.

\noindent\textbf{Random Sampling (RS)} (\cref{fig:retrieve} (a)). We obey the uniform distribution to randomly sample $n$-shot triplets from $\mathcal{D}$.

\noindent\textbf{Retrieving via Similar Image (SI)} (\cref{fig:retrieve} (b)). 
We retrieve $n$ images from $\mathcal{D}$ which are most similar to the query image and then use the corresponding triplets of these retrieved images as the demonstrations. For example, given the query sample $(\hat{\bm{I}},\hat{\bm{Q}})$, suppose the $i$-th image $\bm{I}_i$ is similar to $\hat{\bm{I}}$, then the whole $i$-th triplet $(\bm{I}_i,\bm{Q}_i,\bm{A}_i)$ will be used as one demonstration. Here we use the CLIP embeddings of the images to calculate the cosine similarity.

\noindent\textbf{Retrieving via Similar Texts}. Besides retrieving via images, we can also retrieve $n$ triplets which contain most similar texts to the query sample, where the CLIP embeddings of these texts are used to calculate the cosine similarity. We consider three kinds of texts.

\noindent (1) \textbf{Retrieving via Similar Questions (SQ)} (\cref{fig:retrieve} (c)). We use the question as the text for retrieving, \ie, comparing the similarity between $\hat{\bm{Q}}$ and each $\bm{Q}_i \in \mathcal{D}$.

\noindent (2) \textbf{Retrieving via Similar Question\&Answer (SQA)} (\cref{fig:retrieve} (d)). We concatenate the question and answer into a text sequence for retrieving, \ie,  comparing the similarity between ($\hat{\bm{Q}},\hat{\bm{A}}$) and each $(\bm{Q}_i,\bm{A}_i) \in \mathcal{D}$. Although this strategy cannot be applied in practice since we do not have the ground-truth answer of the query sample, it can give us an ``upper-bound result" of diverse retrieval methods that can help us better analyze other retrieval strategies.

\noindent (3) \noindent\textbf{Retrieving via Similar Question\&Pseudo Answer (SQPA)} (\cref{fig:retrieve} (e)).
Since the ground truth answer is not available during inference, we cannot implement SQA in practice. To exploit the knowledge of the answers in $\mathcal{D}$, we generate the pseudo answer $\hat{\bm{A}}_i^P$ and then concatenate it with $\hat{\bm{Q}}$ for retrieving. To get $\hat{\bm{A}}_i^P$, we can apply the ICL with the demonstrations retrieved by RS and SI.

\subsection{Manipulating Demonstrations}
\label{sec:manipulate_demonstrations}


\noindent\textbf{Mismatching the Triplet}.
To explore whether the correctness of demonstrations affects results, we implement mismatched configurations for the image, answer, and question-answer pair in each demonstration. The following $\Tilde{\bm{I}}$, $\Tilde{\bm{Q}}$, $\Tilde{\bm{A}}$ respectively denotes the mismatched images, questions, and answers.

\noindent (1) \textbf{Mismatching Image (MI)}. We replace the image with a random one from from $\mathcal{D}$. Consequently, $\mathcal{S}$ is transformed to $\{ 
(\Tilde{\bm{I}_1},\bm{Q}_1,\bm{A}_1);...;
(\Tilde{\bm{I}_n},\bm{Q}_n,\bm{A}_n);
(\hat{\bm{I}},\hat{\bm{Q}})\}$.

\noindent (2) \textbf{Mismatching Answer (MA)}. We replace the answer with a random answer in the same label space. $\mathcal{S}$ is transformed to $\{ 
(\bm{I}_1,\bm{Q}_1,\Tilde{\bm{A}_1});...;
(\bm{I}_n,\bm{Q}_n,\Tilde{\bm{A}_n});
(\hat{\bm{I}},\hat{\bm{Q}})\}$.

\noindent (3) \textbf{Mismatching Question-Answer pair (MQA)}. We replace the question-answer pair with a random pair from $\mathcal{D}$. $\mathcal{S'}=\{ 
(\bm{I}_1,\Tilde{\bm{Q}_1},\Tilde{\bm{A}_1});...;
(\bm{I}_n,\Tilde{\bm{Q}_n},\Tilde{\bm{A}_n});
(\hat{\bm{I}},\hat{\bm{Q}})\}$. 

\noindent\textbf{Reordering in Another Modality}.
We reorder the demonstrations based on the similarity of another modality, ensuring that the final sequence is visually and linguistically similar to the query sample.

\noindent (1) \textbf{Reordering SI demonstrations via question similarity (SI-Q)}. 
We use SI to retrieve the demonstrations, and then reorder these demonstrations based on the similarity between the question of each demonstration and $\hat{\bm{Q}}$.

\noindent (2) \textbf{Reordering SQ demonstrations via image similarity  (SQ-I)}.
Similar to SQ-I, we start with SQ to get the initial demonstrations based on linguistically relevance, then reorder the demonstrations by image-lead similarity.

\noindent\textbf{Using Instructions}. 
To investigate how the model behaves when given a specific instruction, we add an instruction at the beginning of the in-context sequence $\mathcal{S'}=\{ 
\bm{Inst};
(\bm{I}_1,\bm{Q}_1,\bm{A}_1);...;
(\bm{I}_n,\bm{Q}_n,\bm{A}_n);
(\hat{\bm{I}},\hat{\bm{Q}})\}$, where $\bm{Inst}$ denotes the instruction. Besides using instructions written by humans, we utilize instructions prompted from GPT-4 to further guide the LVLM.

%% file: sec/4_experiment.tex
\section{Experiments}
\label{sec:experiment}

\subsection{Datasets and Implementation Details}

We utilize three VQA datasets: VQAv2~\cite{goyal2017vqav2}, VizWiz~\cite{bigham2010vizwiz}, and OK-VQA~\cite{marino2019okvqa}. The VQAv2 dataset consists of images from the MSCOCO dataset~\cite{lin2014mscoco}, with more conventional questions. The VizWiz dataset contains low-quality images and questions, and it includes a significant number of unanswerable questions. The OK-VQA dataset requires external knowledge to answer the questions.
In each VQA dataset, we use the training set as our supporting set for the experiments, and the validation set serves as our query set. 
We employ the Open-Flamingo v1(OFv1, the first version of OF) and v2(OFv2, the second version of OF) as the LVLM to evaluate the strategies of demonstration configurations. During retriving, to calculate the embedding similarity, we use the ViT-B/32 model as the vision encoder and a 12-layer Transformer from a well-trained CLIP model~\cite{radford2021clip} as the language encoder to extract image and sentence embeddings, respectively. In ICL, we use 4, 8, 16-shot demonstrations. All experiments are conducted on the RTX 3090 GPU with FP16 precision.

\subsection{Results and Analyses}
Since exhaustive retrieval and manipulation techniques are applied to configure in-context sequence, presenting all these results could lead to disarray. To avoid confusions and emphasize the major conclusions, we show the corresponding experiment results of each claim in the following and list all the results in the Supplementary Material. Next, in Section~\ref{sec 4.2.1}, we first show the inner properties of the applied LVLM, Open-Flamingo (OF), which are concluded from the experiment observations. In this part, we will especially show the limitations of OF that will harm the ICL performance of VQA. Then in Section~\ref{sec 4.2.2}, we will show which configuration strategies can be used to alleviate these limitations for improving the performance.

\subsubsection{The Properties of Open-Flamingo\label{sec 4.2.1}}
In our demonstration configuration experiments and a series of auxiliary experiments, we observe three main properties of OF. Although these properties are specifically observed in OF, which is currently the most suitable LVLM for ICL, the methods used to observe these properties can be applied to all LVLMs. These properties provide new perspectives for interpreting and evaluating the ICL capabilities of LVLMs.

\begin{figure*}[htb]
  \centering
  \includegraphics[width=15.5cm]{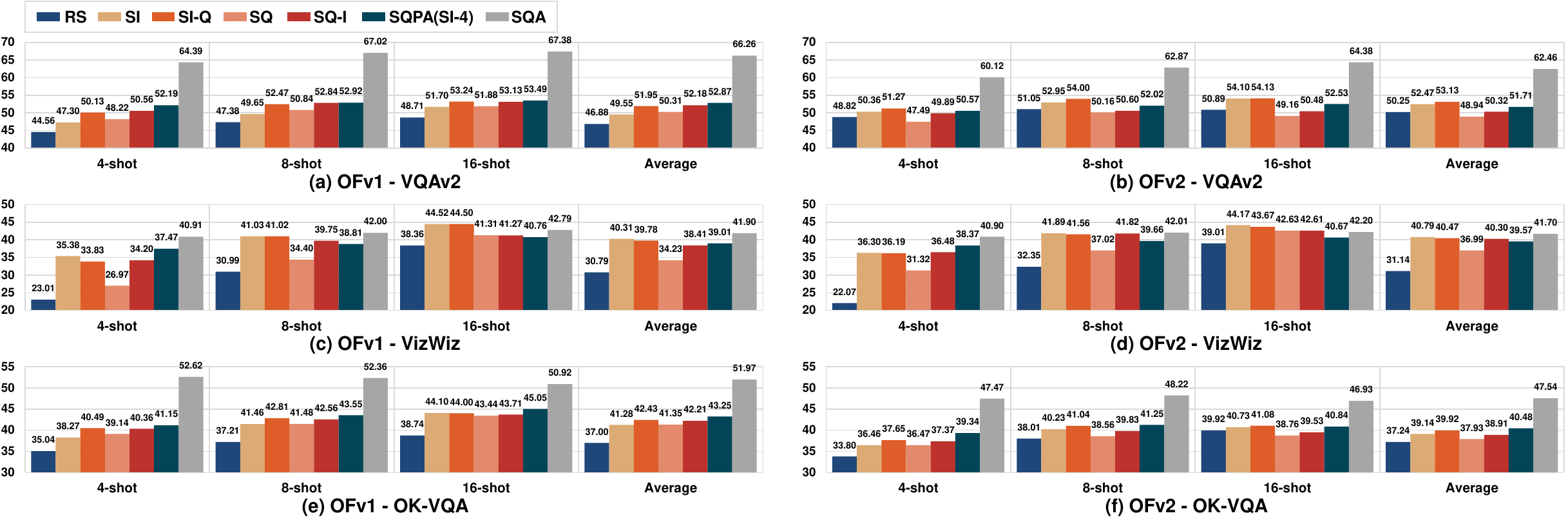}
  \vspace{-0.05in}
  \caption{Experimental results of different demonstrations retrieval strategies on OFv1 and OFv2, comparing 4-shot, 8-shot, 16-shot, and average results across these configurations. SQPA(SI-4) refers to using the result of 4-shot SI as the pseudo answer.}
  \label{fig:icl_result}
  \vspace{-0.2in}
\end{figure*}

\noindent\textbf{Task Recognition (TR) is More Crucial than Task Learning (TL).} This is the first property about OF supported by two experiment observations. Firstly, from \cref{fig:icl_result} we observe that when shot number increases, the accuracy does not consistently increase. For instance, in VQAv2, expanding the shot count from 8 to 16 offers a modest accuracy increase of 1.33 points, compared to a 2.82 point rise from 4-shot to 8-shot(OFv1-RS). 
This suggests that TR outperforms TL in OF, aligning with prior findings that format TR does not significantly benefit from additional shots~\cite{tr&tl}. This is because TR focuses on the label space, format, and input distribution, which means that more shots bring negligible benefits.
However, TL aims at learning input-output mappings. Then given more demonstrations the model can better grasp the mapping relationships and thus enhancing TL performance.

\begin{table}
  \centering
  \scalebox{0.7}{
  \setlength{\tabcolsep}{4mm}{
  \begin{tabular}{@{}l c c  c c c c@{}}
    \toprule
       & \multicolumn{2}{c}{VQAv2} & \multicolumn{2}{c}{VizWiz} & \multicolumn{2}{c}{OK-VQA} \\
       \cmidrule(r){2-3}  \cmidrule(r){4-5} \cmidrule(r){6-7}
       & OFv1 & OFv2 & OFv1 & OFv2 & OFv1 & OFv2 \\
    \midrule
    RS       
    & $45.97$ & $49.94$ & $27.00$ & $27.21$ & $36.13$ &$36.68$\\
    RS(MI)    
    & $45.13$ & $49.73$ & $26.92$ & $35.12$ & $36.41$ &$36.16$ \\
    RS(MA)     
    & $45.65$ & $48.94$ & $12.77$ & $15.03$ & $35.56$ &$29.88$ \\
    \midrule
    SI              
    & $48.48$ & $51.66$ & $38.21$ & $39.10$ & $39.87$ &$38.35$ \\
    SI(MQA) 
    & $47.76$ & $49.94$ & $27.57$ & $26.73$ & $37.37$ &$35.82$ \\
    SI(MA) 
    & $47.64$ & $50.40$ & $13.11$ & $11.48$ & $34.70$ &$29.55$ \\
    \midrule
    SQ 
    & $49.53$ & $48.83$ & $30.69$ & $34.17$ & $40.31$ &$37.52$ \\
    SQ(MI)  
    & $48.01$ & $46.21$ & $30.57$ & $31.95$ & $38.38$ &$33.25$ \\
    SQ(MQA) 
    & $45.32$ & $48.98$ & $27.19$ & $26.45$ & $36.71$ &$35.52$ \\
    SQ(MA)  
    & $47.18$ & $42.72$ & $14.82$ & $15.42$ & $29.50$ &$20.50$ \\
    \midrule
    SQA       
    & $65.71$ & $61.50$ & $41.46$ & $41.46$ & $52.49$ & $47.85$ \\
    SQA(MI)    
    & $65.52$ & $60.88$ & $41.09$ & $40.66$ & $51.31$ & $46.88$ \\
    SQA(MQ)             
    & $50.62$ & $60.62$ & $40.31$ & $40.16$ & $40.25$ & $33.04$ \\
    \bottomrule
  \end{tabular}}
  }
  \vspace{-0.05in}
  \caption{Average results over 4\&8-shot of mismatching triplets.}
  \label{table:mismatch}
  \vspace{-0.2in}
\end{table}

Secondly, Table~\ref{table:mismatch} presents the results of using mismatching triplets. When the disturbed triplets are used, anti-intuitively, the VQA performance does not significantly degrade, \eg, even when all input-output mappings are disturbed, the RS accuracy on VQAv2 only decreases by less than 1 point.
Such phenomenon can be explained from the perspectives of TR and TL. Specifically, TR focus on recognizing the question format, input distribution, and label space from the demonstrations, which can be provided from the disturbed demonstrations as the non-disturbed ones. TL needs to capture the correct input-out mapping while the disturbed QA pairs will damage this. Therefore, TR is less affected by  disturbances compared to TL. 
As the ICL performance is less affected by disturbances, we can conclude that TR plays is more crucial than than TL.

\begin{figure}[t]
  \centering
  \includegraphics[width=7cm]{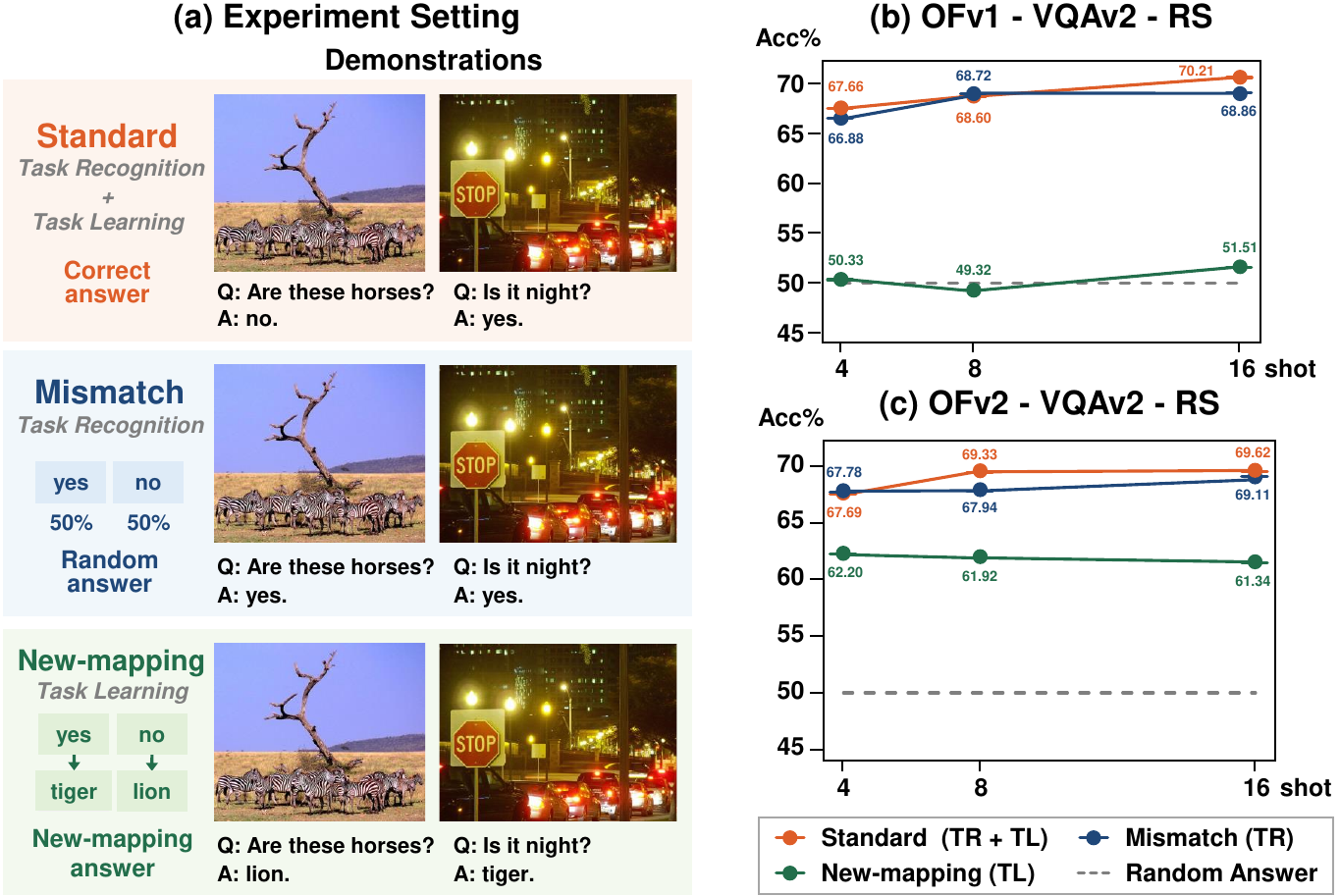}
  \vspace{-0.05in}
  \caption{The evaluation of the TR and TL abilities. Three settings are conducted on the data of VQAv2 whose answer types follow ``yes/no" format: ``Standard" provides the correct demonstrations, preserving the TR and TL capabilities. ``Mismatch" randomly replaces answers to evaluate TR capability. ``New mapping" substitutes ``yes/no" with novel answers (e.g., ``tiger/lion") to test TL capability. The results are shown in (b) and (c).}
  \label{fig:experiment_trtl}
  \vspace{-0.2in}
\end{figure}

Besides these two experiment observations, 
we conduct further validation following the approach proposed in~\cite{tr&tl}. This approach disentangles TR and TL by utilizing different demonstration settings to reflect their respective capabilities. Details of our experiments can be found in the supplementary material.
\cref{fig:experiment_trtl} shows the results, we discover that the accuracy of TR is significantly higher than TL and it is comparable to the results obtained from the standard ICL. Such observation further confirms that TR plays a dominant role in ICL. Additionally, we observe that with more data pre-training and an improved language backbone, the TL capability of OFv2 significantly increases compared to OFv1. This indicates that increasing the amount of pre-training can enhance the TL capability of the model.

\noindent\textbf{Short-cut Inference.}
In NLP, using similar text demonstrations often enhances performance. However, our experiments show that similar demonstrations do not always improve results and can sometimes damage them. For instance, in~\cref{fig:icl_result}, on VQAv2, using demonstrations with similar questions (SQ) performs worse than randomly sampled ones (RS), \eg, 50.25 vs. 48.94. 

\begin{figure*}[htb]
  \centering
  \includegraphics[width=16.5cm]{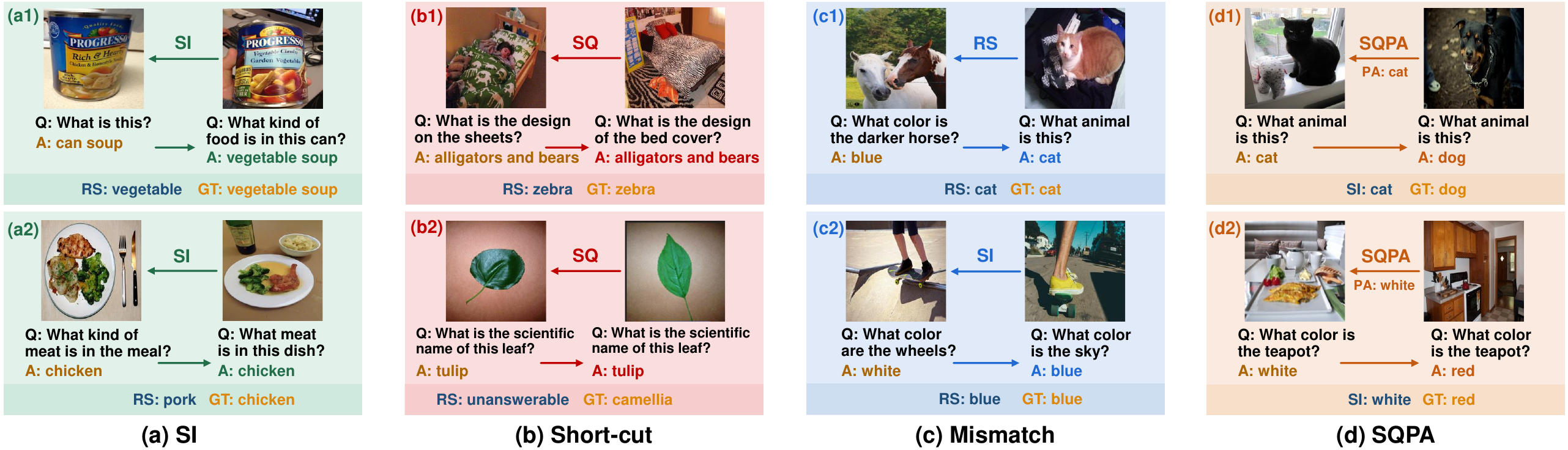}
  \vspace{-0.05in}
  \caption{Four in-context learning scenarios. The arrow above signifies the retrieval strategy, between a retrieved demonstration (left) and the query sample (right). The dark area below indicates the answers to other strategies. (a) \textbf{SI} shows effective TL through similar image retrieval, such as learning ``soup can" in (a1). (b) \textbf{Short-cut} illustrates inference errors where the model replicates the incorrect answer ``tulip" from a similar question in (b2). (c) \textbf{Mismatch} highlights that mismatched answers within the same label space do not significantly impact model effectiveness, with the TR correctly identifying the answer as ``cat" in (c1). (d) \textbf{SQPA} reveals how the model corrects a wrong answer (``white") to ``red" by using pseudo answers for learning the input-output mapping in (d2).}
  \label{fig:visualized}
  \vspace{-0.2in}
\end{figure*}

We believe this happens because \textit{OF tends to build short-cut for predicting}. After analyzing, we find that when the demonstration has a question similar to the query, OF often copies the answer from the demonstration with the similar question instead of using visual information, thus building a short-cut. For example, in \cref{fig:visualized} (b1), when asked about bed sheet patterns, SQ incorrectly returns ``alligator and bear" from the demonstration with a similar question, even though it does not match the query image.

Besides the qualitative observations, we also quantitatively measure the short-cut effect. In \cref{table:copying rate}, we compute the probability that predicted answers also appear in the demonstrations. For SQ, OFv1/OFv2 exhibit copy rates of 77.26\%/79.84\%, respectively, while SQA further increases the copy rates to 87.74\%/89.47\%, while RS and SI achieve only 43.64\%/37.34\% and 50.44\%/54.38\%. Moreover, we conduct an experiment following~\cite{lyu2022copyingeffect} that using demonstrations with identical test inputs and correct or incorrect labels. OF predicts the same answer as the identical input in 47.39\%/45.82\% of cases with correct labels and 37.07\%/45.71\% of cases with incorrect labels, suggesting that even when there is only one question similar to the query in the demonstration, it can still trigger more severe short-cut inference.

This short-cut effect, prevalent in NLP~\cite{lyu2022copyingeffect} and Image Captioning~\cite{wylnips}, its influence beyond LLMs to also impact LVLMs across various tasks. One possible reason is that these models have limited TL ability and are influenced by biases instead of learning from the demonstrations.



\begin{table}
  \centering
  \vspace{0.1in}
  \scalebox{0.7}{
  \begin{tabular}{@{}l c c c c c c@{}}
    \toprule
       & RS & SI & SQ & SQA & SQA(sole) & SQA(sole wrong)\\
    \midrule
    OFv1 & $43.64$ & $50.44$ & $77.26$ & $87.74$ & $47.39$ & $37.07$\\
    OFv2 & $37.34$ & $54.38$ & $79.84$ & $89.47$ & $45.82$ & $45.71$ \\
    \bottomrule
  \end{tabular}
  }
  \vspace{-0.05in}
  \caption{The copy rate (\%) of short-cut on VQAv2 (16-shot). }
  \label{table:copying rate}
  \vspace{-0.2 in}
\end{table}

\noindent\textbf{Image and Language Decoders are not totally Compatible.}
This is the third conclusion about OF and can be demonstrated in two aspects. First, \textit{the language encoder is much stronger than the vision encoder, which causing that linguistic TR plays a more substantial role than visual TR in VQA}. Second, \textit{ the vision and language modules are not aligned well, causing some language reasoning ability lose efficacy in the VL case}.

\begin{figure}[t]
  \centering
  \includegraphics[width=7cm]{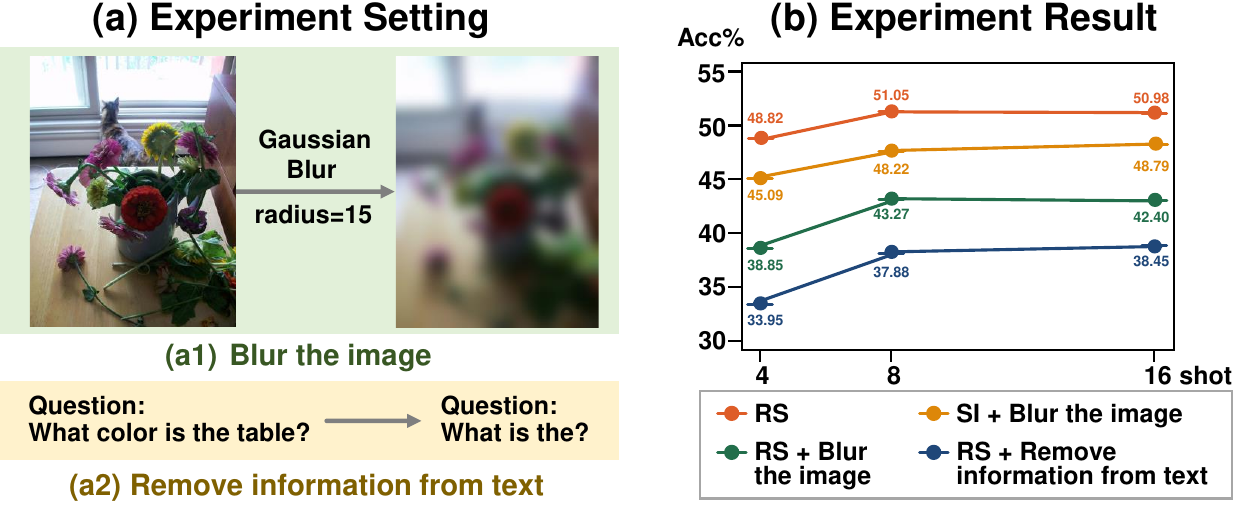}
  \vspace{-0.05in}
  \caption{We add noise to the query image and question: (a1) using Gaussian Blur to blur the image, and (a2) removing information from the question. The experimental results are shown in (b).}
  \label{fig:experiment_noisy}
  \vspace{-0.2in}
\end{figure}

For the first aspect, 
OF shows heightened sensitivity to text quality, with compromised textual input leading to a more significant decline in performance.  
Results in \cref{table:mismatch} demonstrate that replacing the answer in SQ causes a significant 5-point drop, while replacing the image only leads to a minor 1-point decrease.
Similarly, substituting the question of SQA (MQ) lead to a notable decline of 15 points, whereas replacing the image (MI) did not cause a significant decrease.  
Moreover, from \cref{fig:experiment_noisy}, replacing the text with noisy text leads to a more significant performance decline compared to replacing the image with a blurred image. 
These findings validate that linguistic TR plays a more substantial role than visual TR, potentially due to the greater power and scale of language module compared to the visual module in OF, \eg, the language module is LLaMA/MPT, which containing 7 billion parameters and pre-trained on one trillion tokens, while the visual module is CLIP ViT/L-14, containing 428 million parameters and pre-trained on 400 million data.
This indicates that in VLMs, language and vision do not play equally important roles. Instead, linguistic TR demonstrates greater potency and exerts a stronger influence on overall performance.

For the second aspect, we find that some useful strategies for solving QA lose their efficacy in OF. 
For instance, reformulating a QA pair into a declarative sentence to better adapt to the pre-training language model and changing the orders of demonstrations, known to improve performance in NLP, fails to have the same effect on LVLM and detailed experimental descriptions and complete results will be presented in the supplementary materials.  
Additionally, adding instructions before the in-context sequence, which is effective in NLP, only works in OFv2 and not OFv1.  
Before we think that the language module of OF is stronger than the vision module, then why some useful NLP strategies lose the efficacy?
We think the major reason is that the vision and language modules are not aligned well, \ie, the language reasoning of the original LLM does not totally inherited into the VLM after vision-language alignment fine-tuning. Such assumption can be supported from the comparison between OFv1 and OFv2. Compared with OFv1, OFv2 uses more image-text pairs for aligning vision and language modules(180M vs.15M pairs) and thus can inherit more language reasoning ability for solving VL task, and thus we find that adding instructions works better in OFv2 than OFv1, where more details are given in Section~\ref{sec: instruction}.


\subsubsection{Effective Configuration Strategies\label{sec 4.2.2}}
Although OF is one of the SOTA LVLMs for ICL, in section~\ref{sec 4.2.1}, we observe that it has three major limitations: weak TL capabilities, the short-cut effect, and not totally compatible vision-language modules. However, in this section, we still observe that some strategies can improve the ICL ability for VQA.


\noindent\textbf{Similar images and texts lead to better performance.}
Despite we previously show that using demonstrations with similar questions leads to short-cut inference, we now present evidence that using demonstrations that simultaneously contain similar images and questions can enhance performance.
Although the improvements vary depending on the dataset, such strategy is still a powerful way to improve the performance.

First, as \cref{fig:icl_result} shows, using the demonstrations with similar image (SI) consistently boosts LVLM performance, \eg, on VQAv2/VizWiz/OK-VQA/(OFv2, 4-shot), we observe 1.54/14.23/2.66 point improvements. We assume that more similar images in the in-context sequence can compensate more visual information that may have been missed or incorrectly recognized during the visual TR stage. For instance, in \cref{fig:visualized}, while both RS and SQ could only recognize the term ``vegetable" for an image of a soup can, SI identify it as a ``Progresso vegetable soup" since one in-context image also has this soup can. Such visual compensation works more obvious on VizWiz since the image quality of this dataset is quite low and using similar images help OF pinpoint a more accurate label space, \ie, enhancing the visual TR ability.

Secondly, the SQ approach also brings improvements, although these enhancements are not consistently stable due to the presence of the short-cut. However, as shown in \cref{fig:icl_result}, for the VizWiz dataset with lower-quality text and the OK-VQA dataset requiring additional knowledge, demonstrations containing similar questions and reference answers still assist the model in finding the correct answers.

Thirdly, in section~\ref{sec:manipulate_demonstrations}, we show how to reorder the retrieved demonstrations based on their similarity in another modality, \ie, it retrieves similar images/questions and rearranges them based on the similarity of their associated questions/images.
As \cref{fig:icl_result} shows, when applied to two versions of OF and across three varied datasets, this method consistently showcased superiority over base methods. Such findings suggest that both visually and textually similar in-context examples can greatly enhance the performance of LVLMs in TL.

\begin{table}
  \centering
  \scalebox{0.6}{
  \setlength{\tabcolsep}{6.5mm}{
  \begin{tabular}{l l c c c c}
    \toprule
     & Dataset  & 4-shot & 8-shot &   16-shot \\
    \midrule
    RS(OFv1) & VQAv2 & $44.56$ & $47.38$ & $48.71$  \\
           
    Instruct1(OFv1) & VQAv2 & $43.75$ & $46.91$ & $48.67$ \\

    RS & VQAv2 & $48.82$ & $51.05$ & $50.89$  \\

    Instruct1 & VQAv2 & $\textbf{49.93}$ & $\textbf{52.71}$ & $\textbf{50.95}$ \\
    \midrule
    RS & VizWiz  & $22.07$ & $32.35$ & $39.01$   \\

    Instruct1 & VizWiz & $\textbf{25.70}$ & $\textbf{34.71}$ & $\textbf{39.32}$ \\
    \midrule
    RS & OK-VQA & $34.82$ & $38.54$ & $39.55$  \\

    Instruct1 & OK-VQA & $35.72$ & $39.38$ & $40.46$  \\

    Instruct2 & OK-VQA & $\textbf{36.45}$ & $40.17$ & $\textbf{41.11}$  \\

    Instruct3 & OK-VQA  & $35.53$ & $\textbf{40.19}$ & $40.02$ \\
    \midrule
    \multicolumn{5}{p{125mm}}{\textbf{Instruct1}: According to the previous question and answer pair, answer the final question.} \\
    \multicolumn{5}{p{125mm}}{\textbf{Instruct2}: Consider the semantic relationship between the question and the image.} \\
    \multicolumn{5}{p{125mm}}{\textbf{Instruct3}: You will be engaged in a two-phase task. Phase 1: Absorb the information from a series of image-text pairs. Phase 2: Use that context, combined with an upcoming image and your own database of knowledge, to accurately answer a subsequent question.} \\
    \bottomrule
  \end{tabular}}
  }
  \vspace{-0.05in}
  \caption{The results of using instructions.}
  \label{table:instruction}
  \vspace{-0.2in}
\end{table}

\noindent\textbf{Instruction enhances the performance of linguistically advanced model.\label{sec: instruction}}
Providing instructions notably enhances the format TR and TL capabilities of the LVLM. As evident in \cref{table:instruction}, the OFv2 model exhibits substantial improvements across various datasets when using instructions, especially in limited demonstration scenarios. For instance, adding instructions to the 4-shot experiment on VizWiz results in a 3.63 points increment. Given the necessity for additional knowledge in VQA tasks on OK-VQA, we utilize GPT-4 to design two types of instructions: concise and straightforward instructions (Instruct2 in \cref{table:instruction}) and detailed, hierarchical instructions (Instruct3 in \cref{table:instruction}). 
Providing instructions enhances the format TR and TL capabilities of LVLMs by increasing information density in demonstrations, akin to providing more demonstrations. Compared to additional demonstrations, it saves selection time and reduces the processing burden on the visual encoder of LVLM, making it simpler and more convenient.
However, the instructions do not yield significant improvements in experiments with the OFv1 model due to the inferior language encoder of the v1 model, impacting its capability to process these instructions.

\noindent\textbf{Pseudo answers have potential for expeditious enhancement of performance.}
From the results in \cref{table:Pseudo answers}, we can observe that at 4-shot, SQPA generally improves performance. Intuitively, as shown in \cref{fig:visualized}, when the first-round model generates an incorrect answer (``cat"), the demonstration obtained through SQA retrieval using the question and the erroneous answer will be dissimilar to the content of the query image (which is actually a dog). This provides the model with an opportunity to discover that ``cat" is not the correct answer and to reason and infer a new answer. Therefore, the accuracy of the second-round model using SQPA is expected to surpass that of the first round. However, as the number of shots increases, only on OK-VQA does SQPA still show improvement. This may be because too many incorrect QA pairs interfere with the reasoning process of the model, while OK-VQA requires additional knowledge. By using pseudo-answers to search, the model may be able to find more related knowledge.

\begin{table}
  \centering
  \scalebox{0.7}{
  \setlength{\tabcolsep}{0.7mm}{
  \begin{tabular}{@{}l c c c c c c c  c c c c c @{}}
    \toprule
       & \multicolumn{3}{c}{VQAv2} & \multicolumn{3}{c}{VizWiz} & \multicolumn{3}{c}{OK-VQA} \\
       \cmidrule(r){2-4}  \cmidrule(r){5-7} \cmidrule(r){8-10}
       & 4-shot & 8-shot & 16-shot & 4-shot & 8-shot & 16-shot & 4-shot & 8-shot & 16-shot \\
    \midrule
    RS 
    & $48.82$ & $51.05$ & $50.89$ 
    & $22.07$ & $32.35$ & $39.01$
    & $34.82$ & $38.54$ & $39.55$\\
    SQPA(RS-4)    
    & $49.85$ & $51.03$ & $51.96$
    & $30.02$ & $31.93$ & $34.25$
    & $38.92$ & $41.16$ & $40.06$\\
    SI              
    & $50.36$ & $\textbf{52.95}$ & $\textbf{54.10}$
    & $36.30$ & $\textbf{41.89}$ & $\textbf{44.17}$
    & $36.46$ & $40.23$ & $40.73$\\
    SQPA(SI-4)
    & $\textbf{50.57}$ & $52.02$ & $52.53$
    & $\textbf{38.37}$ & $39.66$ & $40.67$
    & $\textbf{39.34}$ & $\textbf{41.25}$ & $\textbf{40.84}$\\
    \bottomrule
  \end{tabular}
  }}
  \vspace{-0.05in}
  \caption{The results of SQPA on OFv2. SQPA(RS/SI-4) refers to using the result of 4-shot RS/SI as the pseudo answer.}
  \label{table:Pseudo answers}
  \vspace{-0.2in}
\end{table}


%% file: sec/5_conclusion_limitations.tex
\section{Conclusion and Limitations}
In this paper, our focus is to investigate the diverse in-context configurations and delve into the inner properties of LVLMs using VQA as a case study. We design various methods to retrieve and manipulate in-context samples. Through exhaustive experiments, we uncover three important inner properties of the applied LVLMs. Furthermore, we identify the strategies that consistently enhance the performance of ICL VQA. These findings contribute to a deeper understanding of LVLMs and provide valuable insights for optimizing their ICL performance in VQA.

However, due to the limited availability of suitable LVLMs for ICL, we currently only explore demonstration configuration and analyze properties of large models on Open-Flamingo. Nevertheless, as discussed at the beginning of \cref{sec:experiment}, the strategies employed in the demonstration configuration and the methods used to observe inner properties of Open-Flamingo can be applied to all LVLMs with ICL capabilities.
In the future, we plan to validate the effectiveness of our proposed demonstration configuration strategies on a wider range of LVLMs. Additionally, we will analyze and evaluate the capabilities of more LVLMs from the perspectives of the three properties observed in Open-Flamingo in \cref{sec 4.2.1}.


%% file: sec/6_appendix.tex
\begin{figure*}[htb]
  \centering
  \includegraphics[width=17cm]{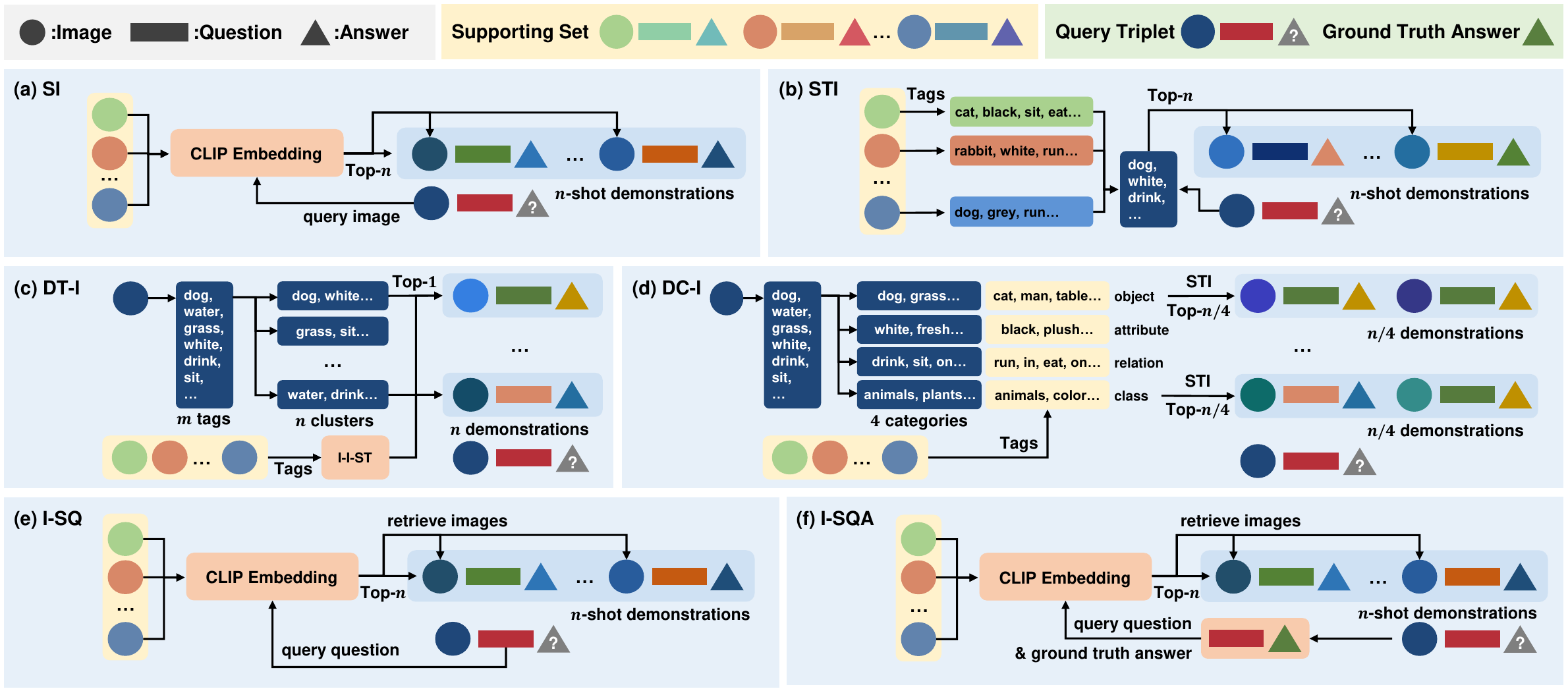}
  \captionof{figure}{The schematic representation of more demonstrations retrieval strategies. (a) is Retrieving via Similar Image (SI) mentioned in Sec.3.2 in the main text of the paper, (b)-(f) is the image retrieval method mentioned in \cref{asec:retrieve_images}. Here we explore more specific retrieval methods, such as focusing on diversity in image retrieval ((c), (d)) and using another modality of information (text) for image retrieval ((e), (f)). }
  \label{fig:a_sti}  
\end{figure*}

\section{More Demonstrations Retrieval Methods}
\label{asec:more_methods}
\subsection{Retrieving Images}
\label{asec:retrieve_images}
Here we introduce more methods centered around retrieving images from $\mathcal{D}$, subsequently using the corresponding triplet as the demonstrations.

\noindent\textbf{(1) Retrieving via Similar Tags from Image (STI)} (\cref{fig:a_sti} (b)).
We employ Vinvl \cite{zhang2021vinvl} and IETrans \cite{zhang2022fine} to extract three categories of tags (object, attribute, and relation) from the images in $\mathcal{D}$ and a given query image. Subsequently, we compute tag overlap between them, aiming to identify and return images from $\mathcal{D}$ exhibiting the highest similarity to the query. Consider a query image tagged with three tags:``dog", ``white" ,``drink". Suppose image A is tagged with ``cat", ``white", ``sit", and image B with ``dog", ``brown", ``drink". STI would prioritize image B due to its higher tag overlap, having two matching tags with the query image. In order to efficiently calculate the overlap in tags, we list all discrete tags utilizing a one-hot manner and then apply an ``AND" operation to assess the similarity at the tag level.

\noindent\textbf{(2) Retrieving via Diverse Image (DI).}
Some works in NLP have discovered that diverse demonstrations containing more relevant information can significantly enhance performance. Drawing on this insight, we retrieve images from $\mathcal{D}$ based on diversity. We extract specific semantic labels from the images and apply two partitioning methods to divide the semantic labels into different clusters to meet the diversity of images for each cluster: \textbf{1) Retrieving via Diverse Tags from Image (DT-I)} (\cref{fig:a_sti} (c)): we first extract three categories of tags (object, attribute, and relation) from $\hat{\bm{I}}$. Suppose there are total $m$ kinds of tags, we divide them into $n$ clusters and there are $m/n$ kinds of tags in each cluster. Within each tag cluster, only the tags in that specific cluster are used to calculate the similarity score between two images, which is based on the number of tags both images share, using the SI method. \textbf{2) Retrieving via Diverse Categories from Image (DC-I)} (\cref{fig:a_sti} (d)): we extract four categories of tags (object, attribute, relation, and class) from the images and categorize the tags into four respective clusters. For example, one cluster will solely contain object tags, while another may only encompass relation tags. This categorization process allows for a more holistic capture of all four content dimensions, retrieving the top-$\bm{n/4}$ similar images within each clusters.

\noindent\textbf{(3) Retrieving Image via Similar Text.} 
In addition to using $\hat{\bm{I}}$ to retrieve images from $\mathcal{D}$, retrieving images based on the query text represents a method that can more effectively leverage the correlation between the visual modality and the linguistic modality. Considering that the image encoder and text encoder of CLIP can respectively map images and text into an embedding space, we employ the CLIP embedding of the query text to retrieve the corresponding embedding of images from $\mathcal{D}$. We propose two methods for text-based image retrieval: \textbf{1) Retrieving Image via Similar Question (I-SQ)} (\cref{fig:a_sti} (e)): we evaluate the similarity between the CLIP embedding of the query question $\hat{\bm{Q}}$ and the CLIP embedding of each image ${\bm{I}}$ in $\mathcal{D}$. \textbf{2) Retrieving Image via Similar Question\&Answer (I-SQA)} (\cref{fig:a_sti} (f)): we use the CLIP embedding of the combination of $\hat{\bm{Q}}$ and the ground truth answer to retrieve images from $\mathcal{D}$ based on similarity metrics.

\subsection{Retrieving Questions and Answers}
\noindent\textbf{(1) Retrieving via Similar Tags from Question (STQ).} We extract different categories of tags from questions. Rather than leveraging the entire question sentence, we select pivotal tags for similarity retrieval. We adopt two types of settings: 1) Use the two most essential tags \textbf{(STQ-2)}: objects and relations. 2) Use four categories of tags \textbf{(STQ-4)}: objects, relations, attributes and interrogative words. Interrogative words are used to identify the type of questions, like ``When".

\noindent\textbf{(2) Retrieving via Diverse Question (DQ).} Similar to DI, more diverse demonstrations enhance the comprehension and reasoning capabilities of the model. In addition to the vision level of diversity retrieval, we also perform diversity retrieval at the language level. We extract four categories of tags from questions: objects, relations, attributes and interrogative words. And we employ STQ to obtain the top-$\bm{n/4}$ similar questions within each category.

\noindent\textbf{(3) Retrieving Text via Similar Image.}
Similar to I-SQ and I-SQA, in addition to using text to retrieve questions from $\mathcal{D}$, we also use images to retrieve text, aiming to combine information from both visual and linguistic modalities. Specifically, we use  $\hat{\bm{I}}$ to retrieve two types of text from $\mathcal{D}$: \textbf{1) Retrieving Question via Similar Image (Q-SI)}: we use the CLIP embedding of $\hat{\bm{I}}$ to retrieve the questions from $\mathcal{D}$. \textbf{2) Retrieving Question\&Answer via Similar Image (QA-SI)}: we use the CLIP embedding of $\hat{\bm{I}}$ to retrieve the question-answer pairs $\{(\bm{Q}_1,\bm{A}_1);(\bm{Q}_2,\bm{A}_2);...;(\bm{Q}_n,\bm{A}_n)\}$ from $\mathcal{D}$.

\subsection{Manipulating Demonstrations}
\noindent\textbf{(1) Changing the Orders of Demonstrations}. 
Some works in NLP have discovered the orders of the demonstrations  can impact performance. Consequently, we also experiment with reversing the order of the demonstrations. We invert the original in-context sequence $\mathcal{S}$ to $\mathcal{S'}=\{(\bm{I}_n,\bm{Q}_n,\bm{A}_n);...;
(\bm{I}_2,\bm{Q}_2,\bm{A}_2)
(\bm{I}_1,\bm{Q}_1,\bm{A}_1);
(\hat{\bm{I}},\hat{\bm{Q}})\}$.

\noindent\textbf{(2) Changing Question-Answer Pairs into Declarative Sentences}. 
Some works like \cite{liu2022declaration} manipulate question-answer pairs into declarative sentences to better adapt to the pre-training language model. We follow \cite{liu2022declaration} to manipulate the question-answer pairs, in which the corresponding short answer is replaced with a [MASK] token. For example, the question ``How many animals are there?" can be changed into ``There are [MASK] animals". Then we add the declarative sentences into the in-context sequence $\mathcal{S'}=\{(\bm{I}_1,\bm{D}_1,\bm{A}_1);...;(\bm{I}_n,\bm{D}_n,\bm{A}_n);(\hat{\bm{I}},\hat{\bm{D}})\}$, where $\bm{D}$ denotes the declarative sentence.


\begin{figure*}[t]
  \centering
  \includegraphics[width=17cm]{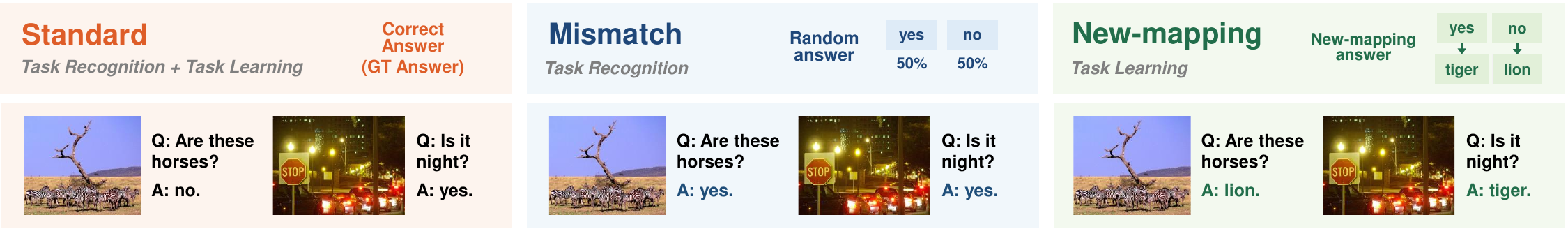}
  \caption{The experiments on disentangling TR and TL. Three settings are conducted: (1) ``Standard" provides the correct demonstrations. (2) ``Mismatch" randomly replaces answers to evaluate TR capability. (3) ``New mapping" substitutes ``yes/no" with novel answers to test TL capability.}
  \label{fig:sm_experiment_trtl}
\end{figure*}

\section{Experiments}
\subsection{More Details about Datasets and Evaluation Metric}
\noindent\textbf{VQAv2.}
VQAv2~\cite{goyal2017vqav2} is a widely-used benchmark for visual question answering, consisting of 443,757 training and 214,354 validation questions. The dataset includes general images sourced from the MSCOCO dataset. Each image is associated with multiple questions and human-annotated answers, reflecting a combination of high-quality visual and commonly encountered questions.

\noindent\textbf{VizWiz.}
The VizWiz dataset~\cite{bigham2010vizwiz} is dedicated to answering visual questions from individuals who are blind. It consists of 20,523 training image/question pairs and 4,319 validation image/question pairs. Blind participants captured images and asked spoken questions, with 10 crowd-sourced answers per question. Due to this, the VizWiz dataset exhibits low-quality images and questions, also with a significant number of unanswerable questions.

\noindent\textbf{OK-VQA.}
OK-VQA~\cite{marino2019okvqa} aims to challenge models to leverage external knowledge for accurate answers. The dataset comprises 14,055 open-ended questions, each associated with five ground truth answers. All questions have been carefully filtered to ensure they necessitate external knowledge, such as information from Wikipedia.

\noindent\textbf{Evaluation Metric.} We follow \cite{alayrac2022flamingo} to use accuracy as the evaluation metric for VQA task. The detailed calculation formula is as follows:
\begin{equation} \label{equ:acc}
    Acc_{a_i} = max(1,\frac{3\times \sum_{k\in [0,9]}^{} match(a_i,g_k)}{10}),
\end{equation}
where $a_i$ denotes the predicted answer of the LVLM, $g_k$ denotes the $k$-th ground true answer, and the $match()$ function indicates whether two answers match, if they match, the result is 1, otherwise it is 0.

\subsection{Implementation Details of Auxiliary Experiments}
\noindent\textbf{Disentangling TR and TL Following ~\cite{tr&tl}.}
We conduct experiments on disentangling TR and TL following ~\cite{tr&tl} as shown in \cref{fig:sm_experiment_trtl}. Specifically, we use different demonstration settings to reflect the TR and TL capabilities of models. We employ the standard in-context learning setting (\textbf{``standard"}) to represent the overall ICL (In-Context Learning) ability of the model, which involves the simultaneous application of TR and TL. Additionally, we use the Mismatching Answer (MA) method (\textbf{``mismatch"}), where the answers in the demonstrations are replaced with random answers that are 50\% correct and 50\% incorrect. This experiment aims to assess the TR ability of model, as format TR only requires the answer space to match the correct answer, while the mapping of incorrect answers significantly affects TL performance. Furthermore, we replace the answer in the demonstrations with an answer from a different answer space, forming a new mapping relationship (\textbf{``new-mapping"}). For example, replacing ``yes" with ``tiger" and ``no" with ``lion". This experiment aims to evaluate the TL ability of model, as such mapping relationships are rarely encountered during pre-training. To facilitate the replacement of the answer space, this experiment is conducted using only the ``yes/no" type questions from the VQAv2 dataset.

\begin{figure}[t]
  \centering
  \includegraphics[width=8cm]{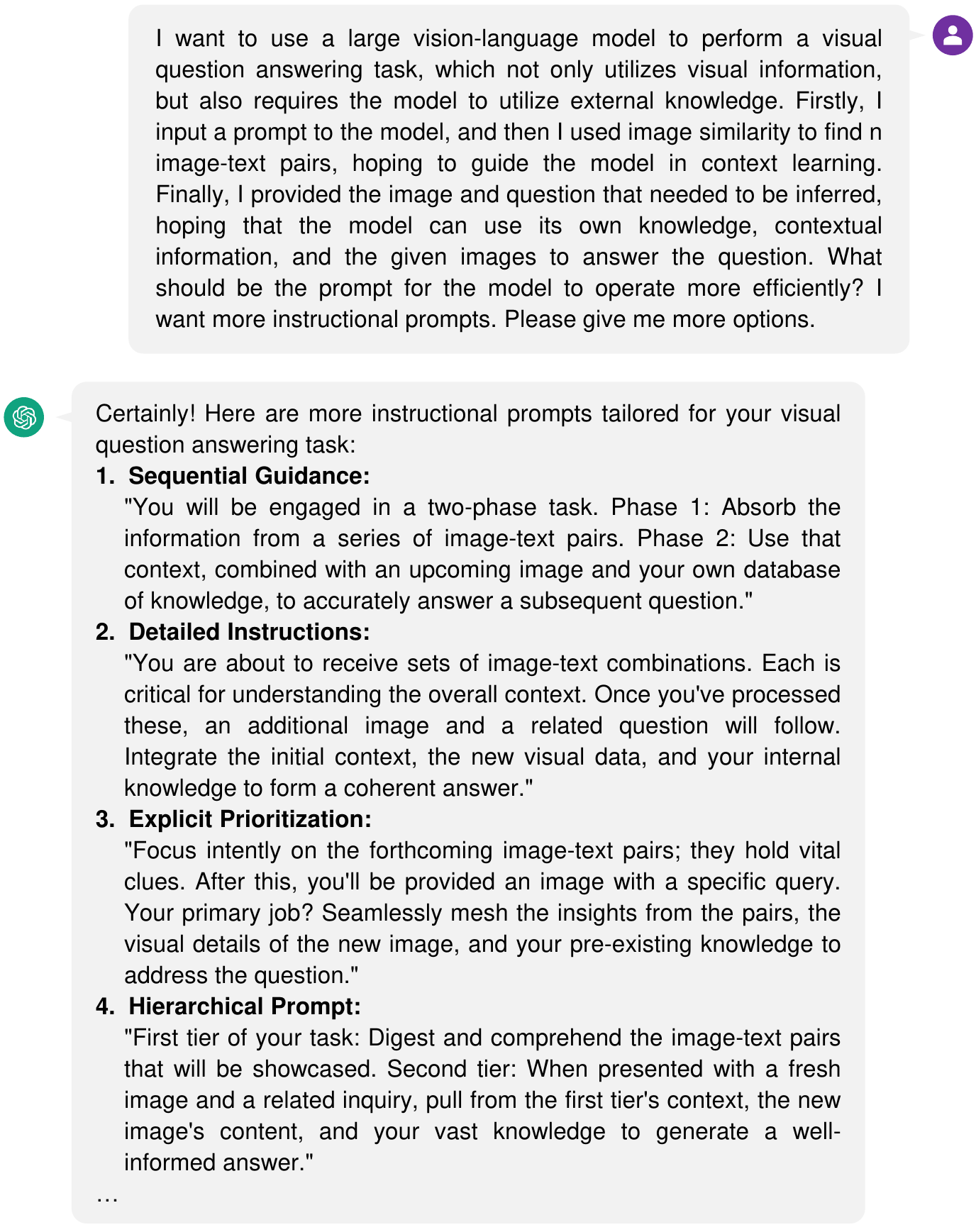}
  \caption{Scenario of using GPT4 for instructions generation . We first use concise language to describe our task, and then inform GPT4 of our requirements.}
  \label{fig:sm_prompt_by_gpt4}
\end{figure}

\begin{figure*}[t]
  \centering
  \includegraphics[width=17.5cm]{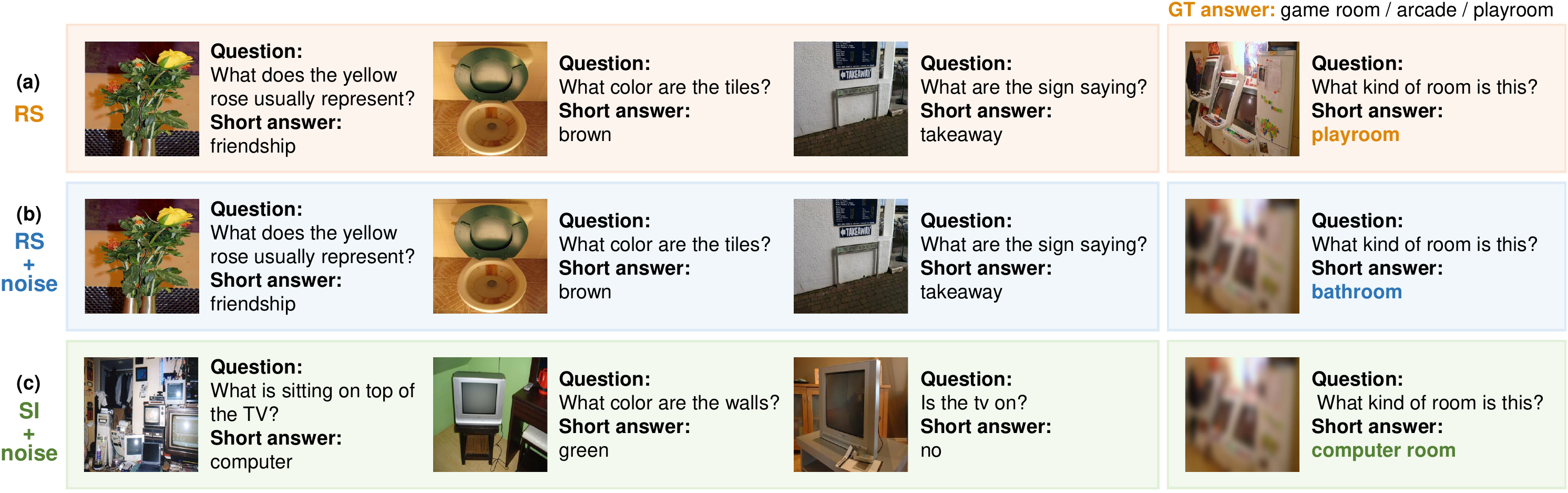}
  \caption{The samples of the experiments on adding noise to the query image. (a) showcases a scenario where demonstrations are randomly sampled and no noise is added to the query image. The model correctly identifies the room as a ``playroom" in response to the question. (b) depicts the use of identical demonstrations but with the query image blurred. The blurred image is difficult to provide effective visual information, resulting in incorrect answers. This error may be influenced by the second demonstration, which erroneously suggests a ``bathroom" scene. (c) employs image similarity for demonstrations retrieval and blurs the query image. However, the demonstrations in this instance provide relevant visual information, enabling the model to identify that this is in a ``computer room".   }
  \label{fig:sm_experiment_noisy}
\end{figure*}

\noindent\textbf{Adding Noise to the Query Image and Question.} 
To better compare the roles of visual information and textual information in TR, we conduct experiments adding noise to the query image and question, effectively diminishing the presence of pertinent information. Specifically, as shown in \cref{fig:sm_experiment_noisy}, we design two distinct experiments: \textbf{1) Adding Noise to the Image}: we apply Gaussian blur to the query image to blur the inherent visual information. As shown in \cref{fig:sm_experiment_noisy} (b) and (c), the noise results in a loss of visual information, leading to errors in answers, while SI method can partially compensate for the loss of visual information. Although the answer is still incorrect in Figure 3 (c), the additional image information enables the model to acknowledge the potential presence of visual elements such as the ``computer" in the original image, which can compensate for the visual TR ability to some extent. \textbf{2) Adding Noise to the Question}: we manipulate the query question by filtering out key information, such as nouns in the question which generally represent the object to be asked. Removing the key information may potentially hinder the model to understand what is being asked.

\subsection{Implementation Details of Preparing Instructions}
To better guide the generation of answers by LVLM, we attempt to add instructions to the input information. Besides using instructions written by humans, we also utilize instructions prompted by GPT-4 to further guide the LVLM. Considering that the VQA task on the OK-VQA dataset relies on external knowledge to answer questions, appropriate instructions might better stimulate the TL capability of the model. Therefore, our primary focus is to explore the impact of diverse instructions on performance for the OK-VQA dataset. Beyond manually written instructions, we have employed plowerful LLM, GPT4\cite{openai2023gpt}, to generate instructions. Specifically, we input our task description and instruction requirements to GPT4, asking it to output various instructions. We mainly require it to generate concise, straightforward instructions and detailed, hierarchical instructions. \cref{fig:sm_prompt_by_gpt4} displays the prompts we use and the responses from GPT-4.

\begin{table*}[htb]
  \centering
  \scalebox{0.8}{
  \setlength{\tabcolsep}{4mm}{
  \begin{tabular}{@{}lcccccc@{}}
  \toprule
          & \multicolumn{3}{c}{OFv1}  & \multicolumn{3}{c}{OFv2}  \\ 
            \cmidrule(r){2-4}  \cmidrule(r){5-7}
       & 4-shot & 8-shot & 16-shot 
       & 4-shot & 8-shot & 16-shot\\
    \midrule
    RS       
    & 44.56  & 47.38  & 48.71   
    & 48.82  & 51.05  & 50.89   \\
    \midrule
    STI
    & 44.61  & 47.89  & 49.91
    & 50.02  & 51.83  & 51.89   \\
    I-SQ      
    & 44.22  & 47.18  & 49.29   
    & 47.14  & 50.47  & 50.96   \\
    I-SQA     
    & 43.84  & 46.58  & 48.31   
    & 47.20  & 48.43  & 49.29   \\
    \midrule
    STQ-2     
    & 48.58  & 50.01  & 51.47   
    & 47.53  & 49.28  & 49.42   \\
    STQ-4     
    & 49.74  & 51.16  & 52.58   
    & 48.69  & 50.18  & 48.83   \\
    Q-SI    
    & 43.07  & 45.56  & 47.33   
    & 46.54  & 49.12  & 48.56   \\
    QA-SI     
    & 45.24  & 47.83  & 49.14   
    & 47.52  & 49.65  & 50.28   \\
    \bottomrule
  \end{tabular}}
  }
  \caption{Experimental Results on the VQAv2 Dataset for more \textbf{Similarity-based Methods} in 4-shot, 8-shot, and 16-shot Learning Settings.}
  \label{table:more_similarity}
\end{table*}

\begin{table*}[htb]
  \centering
  \scalebox{0.8}{
  \setlength{\tabcolsep}{4mm}{
  \begin{tabular}{@{}lcccccc@{}}
  \toprule
          & \multicolumn{3}{c}{OFv1}  & \multicolumn{3}{c}{OFv2}  \\ 
            \cmidrule(r){2-4}  \cmidrule(r){5-7}
       & 4-shot & 8-shot & 16-shot 
       & 4-shot & 8-shot & 16-shot\\
    \midrule
    RS       
    & 44.56  & 47.38  & 48.71   
    & 48.82  & 51.05  & 50.89   \\
    \midrule
    DT-I
    & 46.44  & 48.18  & 49.86
    & 49.27  & 51.75  & 51.09   \\
    DC-I
    & 45.97  & 48.52  & 49.86
    & 48.98  & 51.59  & 50.61   \\
    \midrule
    DQ     
    & 47.03  & 49.24  & 50.24   
    & 49.84  & 51.09  & 49.99   \\
    \bottomrule
  \end{tabular}}
  }
  \caption{Experimental Results on the VQAv2 Dataset for \textbf{Diversity-based Methods} in 4-shot, 8-shot, and 16-shot Learning Settings.}
  \label{table:diversity}
\end{table*}

\begin{table*}[htb]
  \centering
  \scalebox{0.8}{
  \setlength{\tabcolsep}{3mm}{
  \begin{tabular}{@{}lcccccc@{}}
  \toprule
          & \multicolumn{3}{c}{OFv1}  & \multicolumn{3}{c}{OFv2}  \\ 
            \cmidrule(r){2-4}  \cmidrule(r){5-7}
       & 4-shot & 8-shot & 16-shot 
       & 4-shot & 8-shot & 16-shot\\
    \midrule
    RS       
    & 44.56  & 47.38  & 48.71   
    & 48.82  & 51.05  & 50.89   \\
    \midrule
    SI        
    & 47.30  & 49.65  & 51.70
    & 50.36  & 52.95  & 54.1   \\
    SI + Reverse 
    & 47.10  & 49.74  & 51.59
    & 50.54  & 53.23  & 53.84   \\
    \midrule
    SQ        
    & 48.82  & 50.84  & 51.88   
    & 47.49  & 50.16  & 49.16   \\
    SQ + Reverse
    & 47.75  & 49.79  & 50.80
    & 47.38  & 48.77  & 47.74   \\
    \bottomrule
  \end{tabular}}
  }
  \caption{Experimental Results on the VQAv2 Dataset for \textbf{reversing the Order of Demonstrations} in 4-shot, 8-shot, and 16-shot Learning Settings.}
  \label{table:reverse}
\end{table*}

\begin{table}[htb]
  \centering
  \scalebox{0.8}{
  \setlength{\tabcolsep}{3mm}{
  \begin{tabular}{@{}lccc@{}}
  \toprule
    & 4-shot & 8-shot & 16-shot\\
    
    \midrule
    RS       
    & 48.82  & 51.05  & 50.89   \\
    RS + Declarative Sentences       
    & 33.19  & 38.61  & 39.21   \\
    \midrule
    SI*        
    & 51.23  & 52.14  & 52.55   \\
    SI* + Declarative Sentences
    & 41.89  & 44.04  & 44.11   \\
    \bottomrule
  \end{tabular}}
  }
  \caption{Experimental Results on the VQAv2 Dataset for \textbf{changing Question-Answer Pairs into Declarative Sentences} in 4-shot, 8-shot, and 16-shot Learning Settings. SI* denotes that we use SI to retrieve images and limit the retrieved images to be non repetitive.}
  \label{table:declaration}
\end{table}

\section{More Experimental Results}

Here, we present all the numerical results from the experiments, which are based on different demonstration retrieval and manipulation methods that are not mentioned in the main text. These methods are not discussed in the main text because they do not show significant improvement in the performance of OF or exhibit unstable results.

\cref{table:more_similarity} showcases more similarity-based retrieval methods, including tag-based retrieval, which yields similar results to the embedding-based retrieval method mentioned in the main text. Both approaches exhibit stable improvements. However, using mixed modality for retrieval (I-SQ, I-SQA, Q-SI, and QA-SI) does not provide significant assistance to the performance of model and may even lead to performance degradation. This could be attributed to the limited amount of image-related information present in the questions. Consequently, using images to retrieve questions or using questions to retrieve images is not a viable choice.

\cref{table:diversity} displays the results of using diversity-based retrieval methods. It can be observed that enhancing the diversity of similar demonstrations can improve the results to some extent, but compared to similarity retrieval, it cannot yield significant improvements for OF.

\cref{table:reverse} presents the impact of inputting similar demonstrations in reverse order to the model. Previous studies in the field of NLP~\cite{liu2021selection} have found that placing more similar samples closer to the query can lead to greater performance improvements. However, based on the experimental results in this paper, such manipulation has little effect on OF.

\cref{table:declaration} illustrates the effect of changing questions into declarative sentences. This transformation evidently leads to a marked reduction in performance of the model. One potential explanation for this phenomenon could be the difficulty for the model in recognizing the [mask] token. This challenge hinders the model to grasp the underlying reasoning requirements in the task. Consequently, this approach does not yield a significant improvement in OF either.

\cref{table:instruction_sm} presents the outcomes of offering different instructions. Both the concise instructions and the detailed, hierarchical instructions result in noticeable enhancements compared to the Random Sampling (RS) outcomes. Notably, concise and pertinent instructions appear to yield superior results. Building on these findings, we plan to delve deeper into the efficacy of different instructional methodologies.





\section{Experiments on More Large Vision-Language Models}
In our study, we primarily utilized Open-Flamingo, a Large Vision-Language Model (LVLM), as our experimental model. Additionally, we also explored other LVLMs that support in-context learning. For instance, we examine Otter\cite{li2023otter}, a LVLM based on Open-Flamingo, is fine-tuned on the MIMIC-IT multimodal dataset\cite{li2023mimic} for enhancing performance.

Furthermore, We compare the performance of these models trained on different large language models (LLMs). The primary distinction between Otter v1 and Otter v2 lies in their language models, similar to Open-Flamingo, Otter v1 utilizes LLAMA-7B, while Otter v2 employs MPT-7B. The outcomes of these comparative experiments are shown in \cref{table:different_models}.

\begin{table*}[h]
  \centering
  \renewcommand\arraystretch{1.3}
  \tabcolsep=3mm
  \scalebox{0.8}{
  \begin{tabular}{@{}p{10cm}  c c c@{}}
    \toprule
    Instruction  & 4 shot & 8 shot &   16 shot \\
    \midrule
    RS
    & 34.82 & 38.54 & 39.55 \\
    
    According to the previous question and answer pair, answer the final question.
    & 35.72 & 39.38 & 40.46 \\

    Using external knowledge and image content to answer questions.
    & 34.28 & \textbf{40.47} & 40.69 \\

    Integrate information from the question, image, and previous answers.
    & 36.40 & 40.46 & 40.88 \\

    Consider the semantic relationship between the question and the image.
    & \textbf{36.45} & 40.17 & \textbf{41.11} \\

    For the upcoming tasks, you'll be provided image-text pairs. Digest these pairs carefully. Later, an image along with a question will be presented. Combine your understanding from the pairs, the new image, and your own knowledge to answer.
    & 35.13 & 40.30 & 40.61 \\
    
    You will be engaged in a two-phase task. Phase 1: Absorb the information from a series of image-text pairs. Phase 2: Use that context, combined with an upcoming image and your own database of knowledge, to accurately answer a subsequent question.
    & 35.53 & 40.19 & 40.02 \\
    \bottomrule
  \end{tabular}
  }
  \caption{Experimental Results of RS on the VQAv2 Dataset for \textbf{using different instructions} in 4-shot, 8-shot, and 16-shot Learning Settings.}
  \label{table:instruction_sm}
\end{table*}

\begin{table*}[t]
  \centering
  \scalebox{0.8}{
  \setlength{\tabcolsep}{4mm}{
  \begin{tabular}{@{}lllcccc@{}}
  \toprule
      & LVLM & Language Model & 4-shot & 8-shot & 16-shot & Average \\
    \midrule
    RS & Open-Flamingo & LLAMA-7B      
    & 44.56  & 47.38  & 48.71 & 46.88 \\
    RS & Open-Flamingo & MPT-7B   
    & 48.82  & 51.05  & 50.89 & 50.25 \\
    RS & Otter & LLAMA-7B      
    & 39.14  & 41.28  & 42.43 & 41.13 \\
    RS & Otter & MPT-7B   
    & 24.96  & 27.60  & 29.92 & 27.49 \\
    \midrule
    SI & Open-Flamingo & LLAMA-7B       
    & 47.30  & 49.65  & 51.70 & 49.55 \\ 
    SI & Open-Flamingo & MPT-7B
    & 50.36  & 52.95  & 54.10  & 52.47 \\
    SI & Otter & LLAMA-7B      
    & 37.61  & 39.72  & 41.42 & 39.58 \\
    SI & Otter & MPT-7B      
    & 23.08  & 24.38  & 23.95 & 23.80 \\
    \midrule
    SQ & Open-Flamingo & LLAMA-7B       
    & 48.82  & 50.84  & 51.88 & 50.82 \\
    SQ & Open-Flamingo & MPT-7B
    & 47.49  & 50.16  & 49.16 & 48.94 \\
    SQ & Otter & LLAMA-7B      
    & 39.16  & 39.41  & 40.57 & 39.71 \\
    SQ & Otter & MPT-7B      
    & 23.29  & 23.56  & 22.74 & 23.20 \\
    \bottomrule
  \end{tabular}}
  }
  \caption{Experimental Results on the \textbf{different Large Vision-Language Models} in 4-shot, 8-shot, and 16-shot Learning Settings.}
  \label{table:different_models}
\end{table*}

\begin{figure*}[t]
  \centering
  \includegraphics[width=17.5cm]{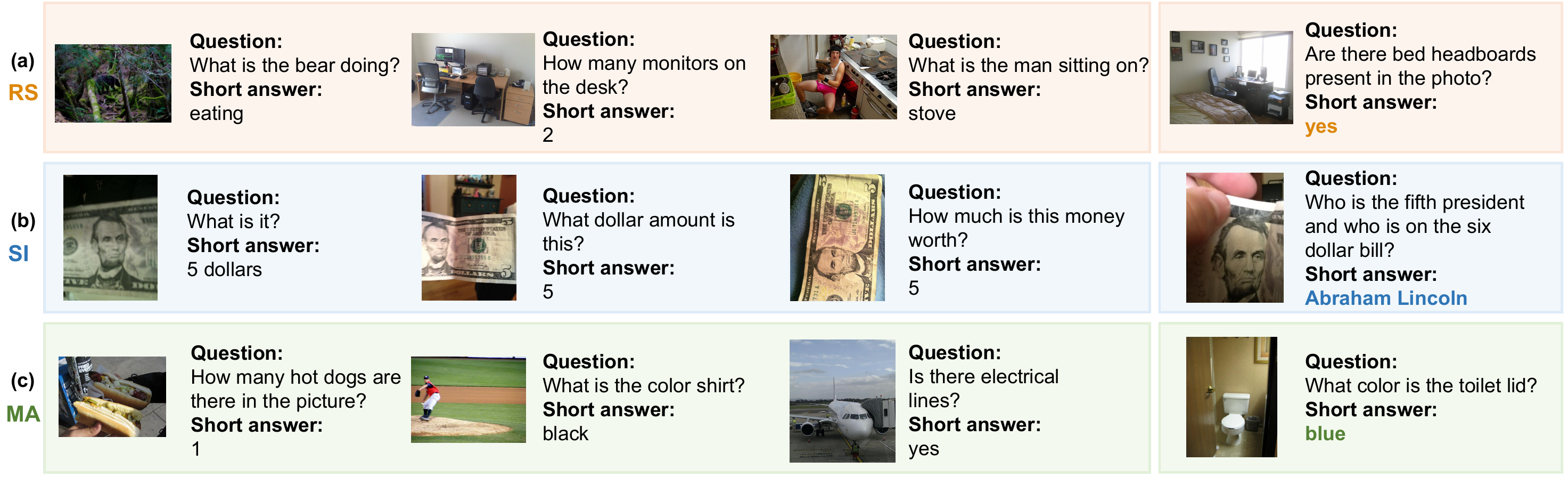}
  \caption{More input-output samples from our experiment of Open-Flamingo.}
  \label{fig:sm_example_cases}
\end{figure*}